\documentclass[letterpaper]{article} 
\usepackage{aaai20}  
\usepackage{times}  
\usepackage{helvet} 
\usepackage{courier}  
\usepackage[hyphens]{url}  
\usepackage{graphicx} 
\usepackage{verbatim}
\usepackage{graphicx}  
\usepackage{booktabs}       
\usepackage{amsfonts}       
\usepackage{nicefrac}       
\usepackage{microtype}      
\usepackage{amsmath}
\usepackage{algorithm}
\usepackage{algorithmic}
\usepackage{mathrsfs,dsfont}
\usepackage{amsthm,amssymb}
\usepackage{lipsum}
\usepackage{bm}

\newtheorem{proposition}{Proposition}

\newcommand{\citep}{\cite}
\newcommand{\citet}[1]{\citeauthor{#1} \shortcite{#1}}
\usepackage{subfigure}
\usepackage{color}
\nocopyright

\definecolor{xinruncolor}{RGB}{138,43,226}

\title{Inducing Cooperation via Team Regret Minimization based Multi-Agent Deep Reinforcement Learning }
\author{Runsheng Yu\textsuperscript{\rm 1$\dagger*$}, Zhenyu Shi\textsuperscript{\rm 2$*$},  Xinrun Wang\textsuperscript{\rm 1}, Rundong Wang\textsuperscript{\rm 1}, Buhong Liu\textsuperscript{\rm 3}, \\
	\Large \textbf{Xinwen Hou\textsuperscript{\rm 4}, Hanjiang Lai\textsuperscript{\rm 2},  Bo An\textsuperscript{\rm 1}}\\
	\Large \textsuperscript{\rm 1}Nanyang Technological University\space\space\textsuperscript{\rm 2}Sun Yat-sen University \space\space\textsuperscript{\rm 3}King's College London\\
	\Large\textsuperscript{\rm 4}Institute of Automation, Chinese Academy of Sciences\\
	\textsuperscript{\rm $\dagger$}yu-jusheng@foxmail.com \\
		\textsuperscript{\rm $*$}These authors contributed equally.
}

\begin{document}
\maketitle

\begin{abstract}

Existing value-factorized based Multi-Agent deep Reinforcement Learning (MARL) approaches are well-performing in various multi-agent cooperative environment under the \emph{centralized training and decentralized execution} (CTDE) scheme, where all agents are trained together by the centralized value network 
and each agent execute its policy independently. However, an issue remains open: in the centralized training process, when the environment for the team is partially observable or non-stationary, i.e., the observation and action information of all the agents cannot represent the global states, existing methods perform poorly and sample inefficiently. 
Regret Minimization (RM) can be a promising approach as it performs well in partially observable and fully competitive settings. However, it tends to model others as opponents and thus cannot work well under the CTDE scheme.
In this work, we 
propose a novel team RM based Bayesian MARL with three key contributions: (a) we design a novel RM method to train cooperative agents as a team and obtain a team regret-based policy for that team; (b) we introduce a novel method to decompose the team regret to generate the policy for each agent for decentralized execution; (c) to further improve the performance, we leverage a differential particle filter (a Sequential Monte Carlo
method) network to get an accurate estimation of the state for each agent. 
Experimental results on two-step matrix games (cooperative game)  and battle games (large-scale mixed cooperative-competitive games) demonstrate that our algorithm significantly outperforms state-of-the-art methods.
\end{abstract}

\section{Introduction}

Decentralized cooperative multi-agent systems (DCMAS) have attracted increasing attention with their successful applications to several domains, e.g., robotics, recommender systems and communications~\citep{yang2004multiagent,bikov2015multi,kim2019learning}.
The key problem in DCMAS is how to train agents to cooperate well with each other in the partially observable environment and with \emph{only} team reward, which is referred to as Decentralized Partially Observable Markov Decision Process (Dec-POMDP)~\citep{oliehoek2016concise}. MARL is known as one of the promising methods to solve Dec-POMDPs~\citep{hernandez2018multiagent} and a well-known scheme for MARL is the \emph{centralized training and decentralized execution} scheme, which enables each agent to learn a well-performing policy through centralized training and executes each agent's policy independently in the decentralized execution stage~\citep{lowe2017multi,oliehoek2008optimal}. Based on this scheme, a surge of methods have been designed recently~\citep{sunehag2017value,foerster2018counterfactual,rashid2018qmix,peng2017multiagent}. Among these methods, value-factorized MARL approaches have received much attention. 
Briefly speaking, for the value-factorized MARL, all the cooperative agents are regarded as a team and then, the team is trained through the standard value-based training methods, e.g., Q-learning, using the observation and action information of all the agents as the state input.

Although tremendous progress has been made, the following research question remains open for value-factorized MARL: how to deal with the environment which is \emph{partially observable} or \emph{non-stationary} from the perspective of the \emph{team}? 
The value-based centralized training methods can work well only when the environment is stationary and fully observable for the team~\citep{iqbal2018actor}. 
But when the current observations and actions of all agents, i.e., the team, cannot fully represent the global states, these MARL methods suffer from poor performance and sample inefficiency. This issue is commonly seen when the environment is complicated (e.g., the number of agents is large). Even when a good approximation to the global states is given, the non-stationary environment (some elements change over episodes, e.g., there exist opponents whose strategies might change over time) still degrade the performance. For brevity, we term this environment as the \emph{Team partially Observable or Non-stationary Environment} (TONE). Some scholars attempt to mitigate these issues by adding global states to the team agent~\citep{rashid2018qmix} but these approaches are hard to use in large-scale environment due to the large dimensions of global states and still suffer from the non-stationary environment. 

The regret minimization (RM) is a promising approach to address the poor performance and sample inefficiency challenges under TONE because it maintains an accumulated regret learned from past experience to perceive the change of the environment~\citep{zinkevich2008regret,jin2017regret,srinivasan2018actor}. Therefore, it can be applied to both stationary and non-stationary environment, given that most of the value-based training methods are only suitable in the stationary and fully observable environment. However, it is inappropriate to directly apply RM into Dec-POMDPs with cooperative agents. The primary reason behind that is the RM tends to model any other agents in the environment as opponents~\citep{zinkevich2008regret,jin2017regret,brown2019deep}, if we let each agent learn by RM individually through the team reward\footnote{Some previous researches have explored the individual regret method for planning in Dec-POMDPs but with strong assumptions~\citep{wu2017multi}.} (as shown in Section~\ref{exp}). Therefore, a novel RM framework which is capable to consider cooperation is required.

Moreover, inspired by the fact that the more the agent knows about its current state, the better action it will take, it is possible to introduce a novel tool to infer the current state for each agent to further reduce the uncertainty and improve the performance of each agent. 

\par 

\begin{figure}
\centering
\centerline{\includegraphics[width=\linewidth]{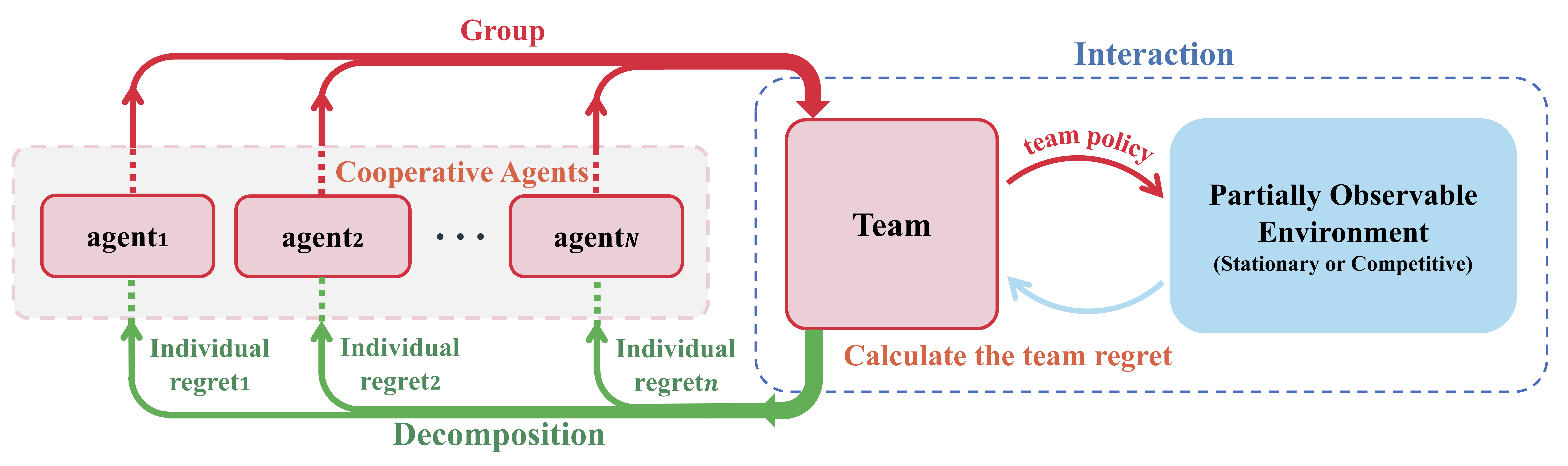}}
\caption{The schematic illustration of the whole framework. The cooperative agents as grouped a \emph{team}. We learn the team regret and decompose it to each agent as individual regret. Details can be found in Section~\ref{Team Regret Minimization}.}\label{fig1}
\end{figure}

\textbf{Contributions}. To solve these problems jointly, in this paper, we introduce a novel team RM based Bayesian MARL framework as shown in Fig.~\ref{fig1}.  The main contributions can be summarized as follows: (a) We design a novel RM method to train cooperative agents as a team and obtain a regret-based team policy which can generate well-coordinated actions in TONE;  (b) we introduce a novel method to decompose the team regret to generate the individual regret based policy for each agent to guarantee that the team policy is the same as the collection of individual policies of each agent for decentralized execution. To our best knowledge, this is the first to combine RM to the centralized training and decentralized execution scheme; (c) we leverage a differential particle filter network, which maintains a belief state to remember the history information and thus aids each agent to make better inference of its current state. 
In our experiments, we evaluate our approaches in the environment of  two-step matrix game (cooperative game) and large-scale battle games (mixed cooperative competitive games). Experiment results demonstrate that the proposed approaches significantly outperform baseline methods.

\section{Related Works}
The first line of related research is the \emph{value-factorized MARL}. One of the simplest MARL methods is independent Q-learning, which only regards other team agents as part of the environment without cooperative effect~\citep{tan1993multi}. 
To consider cooperation, researchers introduce value-factorized methods: 
\citet{sunehag2017value}
design the Value-Decomposition Networks (VDN) to learn cooperation by the linear summation based factorized method, but this approach may not capture the complex interaction between agents. 
\citet{rashid2018qmix} leverage the hyper-network to provide extra global-state information to estimate the value function, but it is time-consuming when the number of agents increases and still does not consider the non-stationary environment. Another approach utilizes an affine transformation function as the decomposition function to provide a more flexible training approach~\citep{son2019qtran}. Although these centralized training methods provide well-performing methods for value decomposition, all of them do not take the TONE into consideration, which may lead to poor performance and sample inefficiency.
Different from other factorized methods, we propose a novel RM based centralized training scheme to alleviate the uncertainty and enhance the agents' performance in TONE.



The second line is the \emph{counter-factual based learning}. These methods are widely used to solve DCMAS, involving counter-factual reward method and counter-factual value method. The counter-factual reward method shapes the reward using domain knowledge for obtaining more accurate contribution of certain agent, but it is hard to generalize~\citep{devlin2014potential}. The counter-factual value approach learns coordination by calculating the counter-factual baseline value of each agent within a team~\citep{foerster2018counterfactual} but this approach focuses on calculating each agent's counter-factual baseline value which is time-consuming in large-scale environment and only considers the instant counter-factual value. Differently, our method provides a time-saving approach to calculate each agent's counter-factual value and also considers the accumulated counter-factual value to perceive the change of the environment.

The third line is \emph{regret minimization based RL method}, which combines the RM and RL together. \citet{jin2017regret} propose an advantage-based regret minimization (ARM) algorithm in POMDP but only consider the single-agent case. \citet{srinivasan2018actor} take advantage of the regret policy gradient in multi-agent environment but they only investigate the fully competitive game, i.e., no cooperative agents are considered. Our method is also a variant of regret minimization based RL methods but differently. To the best of our knowledge, we are the first to leverage RM into the Dec-POMDPs through learning the team regret in centralized learning scheme. 

The forth line is the variational inference based RL. The key idea of these approaches is to leverage variational inference to provide a good inference of the current state. For instance, \citet{igl2018deep} apply a variational autoencoder based RL approach to provide better inference of current belief state distribution in POMDP; \citet{katt2019bayesian} leverage the particle filter method in factorized-POMDP as a current state estimator. 
We take this idea by introducing a novel end-to-end particle filter to estimate each agent's belief distribution in Dec-POMDPs.  

\section{Preliminaries}
In this section, we present some 
fundamental ideas of the Dec-POMDPs as well as the RM based RL methods. 
\subsection{Dec-POMDP} 
\par 
A cooperative multi-agent task\footnote{Our methods can also deal with mixed cooperative and competitive tasks, as shown in Section~\ref{exp}.} can be formally represented as a Dec-POMDP. 
Consider a Dec-POMDP represented as a tuple $\langle \mathcal{S},\mathcal{N},\mathcal{A},P,r,\mathcal{O},\gamma,b_1\rangle$, where $s\in \mathcal{S}$ is the global states; $o^i\in \mathcal{O}^i$ is the own observation for agent $i$ and $\times_{i=1}^N\mathcal{O}^i = \mathcal{O}$; $i\in\mathcal{N}$ is the $i$-th agent and $|\mathcal{N}|=N$; $a^i \in \mathcal{A}^i$ is the agent $i$'s action, $\mathcal{O}^i \rightarrow P(A)$, which maps each agent’s observation to a distribution of actions, and $\mathbf{a}\in \mathcal{A} = \times_{i=1}^N\mathcal{A}^i$ is a joint action; $P(s_{\tau+1}|s_{\tau},\mathbf{a}_\tau)$ is the transition function: $\mathcal{S}\times \mathcal{A} \rightarrow \mathcal{S}$. 
The reward function $r(s,\mathbf{a}): \mathcal{S} \times \mathcal{A} \rightarrow \mathbb{R}$ is used to generate reward given the global states $s\in \mathcal{S}$ and the joint action $\mathbf{a} \in \mathcal{A}$.  
$\gamma$ is the discount factor and $a^i = \pi^i(a^i|o^i)$ is the policy for agent $i$. All the agents share \emph{the same reward}, named as the team reward. $b_1$ is the initial belief state with density $b_1(s_1)$.
\par 
Our goal is to find the optimal joint policy for $N$ agents: $\boldsymbol{\pi}=\begin{bmatrix}\pi^1,\pi^2,\dots,\pi^N\end{bmatrix}$ that can maximize the total reward within a team, formed as an optimization problem: $\arg\max_{\boldsymbol{\pi}} [\sum\nolimits^{\infty}_{\tau=1}\gamma^{\tau}r_{\tau}(s_\tau,\mathbf{a}_\tau)]$, where $\tau$ \emph{is the time step}\footnote{In the case without causing confusion, $r_{\tau}(s_\tau,\mathbf{a}_\tau)$ is hereinafter referred to as $r_{\tau}$.}. 

\subsection{Regret Minimization based RL}


\par 

Advantaged-based Regret Minimization (ARM)~\citep{jin2017regret} is a well-performing RM based RL method in the Zero-Sum Markov Game (ZMG). 

Before discussing some details about ARM, we define some notations here: for agent (player) $i$, let $\mathcal{I}^i$ be the space of information for agent $i$ and $I^i \in \mathcal{I}^i$ is the set of agent $i$'s sequence of observations, i.e., 
the information state for agent $i$ at step $\tau$ can be represented as:  
$I^i_\tau = (b_0,a^i_0,o^i_1,\dots,a^i_{\tau-1},o^i_\tau)$\footnote{$I^i_{\tau}$ is only to stress that at $\tau$-th step, the agent $i$ owns the information state $I^i$.}.  
The primary idea of ARM is to find the accumulated regret of agent $i$: 
$\operatorname{REG}^i_{1:T}(I^i,a^i)=\sum\nolimits_{t=1}^{T}Q^{i}_{\boldsymbol{\pi}_t}(I^i,a^i)-V^{i}_{\boldsymbol{\pi}_t}(I^i)$ through learning,
where $t$ as the $t$-th \emph{episode}\footnote{Since we need to use the notations of both episode and time step, to avoid confusion, we use $t$ to represent $t$-th episode and $\tau$ to represent $\tau$-th step.} within total $T$ episodes and $Q^{i}_{\boldsymbol{\pi}}(I^i,a^i):I\times A\rightarrow \mathbb{R}$ is the counter-factual state-action value of agent $i$, indicating the value (how good or bad) when agent $i$ performs action $a^i$ at information state $I^i$. Value $V^{i}_{\boldsymbol{\pi}}(I^i) = \sum_{a^i\in A}\pi^i(a^i|I^i)Q^{i}_{\boldsymbol{\pi}}(I^i,a^i)$ is the expected value function for agent $i$ of taking joint strategy $\boldsymbol{\pi}$.


\par 
The key idea to estimate the immediate counter-factual value for agent $i$ in ARM at information state $I^i_{\tau}$ is using the reward: $Q^{i}_{\boldsymbol{\pi}}(I_{\tau}^i,a_{\tau}^i) -V^{i}_{\boldsymbol{\pi}}(I_{\tau}^i) = \mathrm{r}_{\tau}^i$, where $\mathrm{r}_{\tau}^i$ is the individual reward for agent $i$ at $\tau$-th step in ZMG. This equation builds a connection between the reward and the counter-factual value and thus the counter-factual value can be learned directly from the temporal difference learning~\citep{sutton1998introduction}. 

\section{Algorithm Framework}

\begin{figure}
\centering
\centerline{\includegraphics[width=0.9\linewidth]{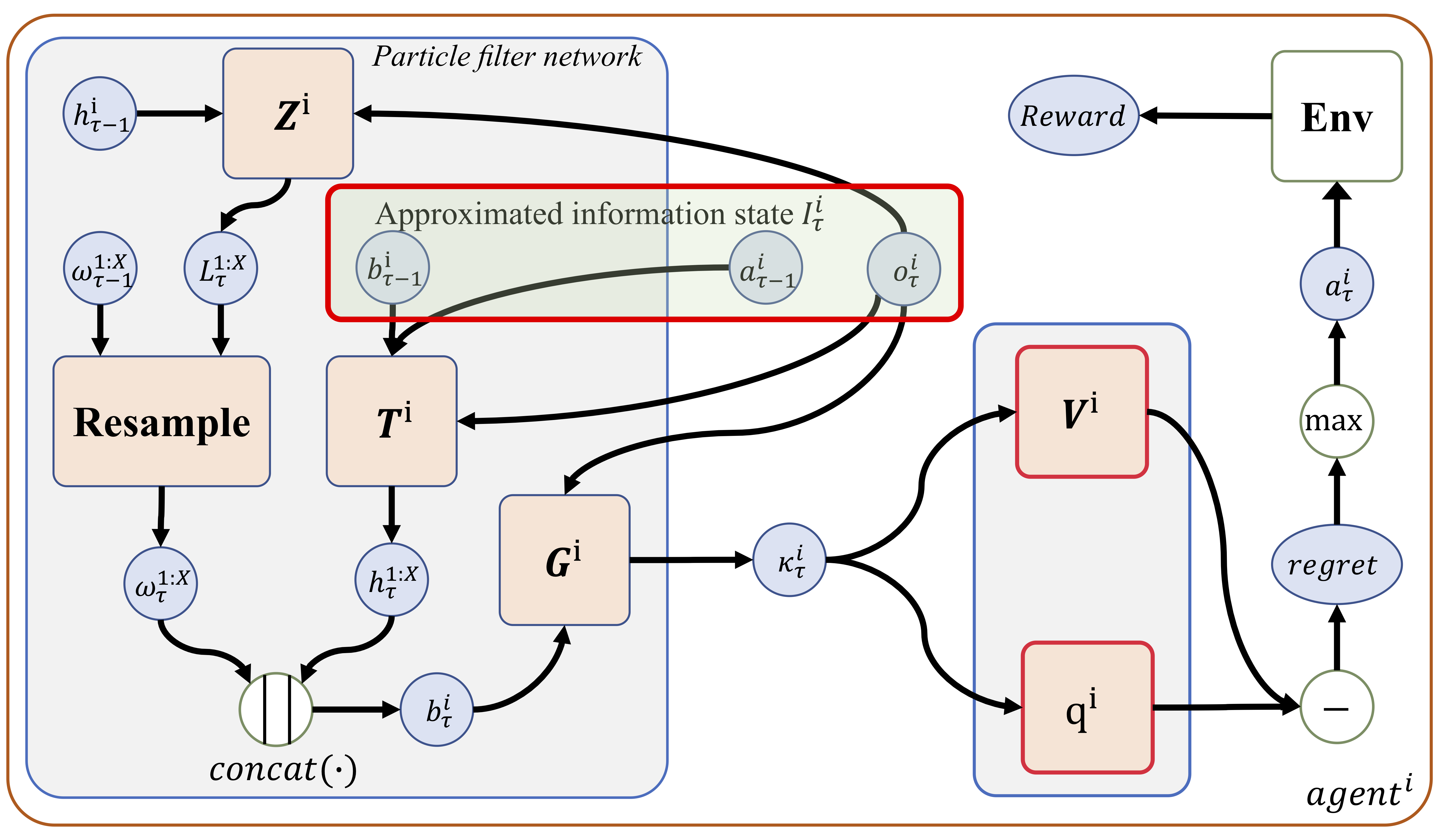}}
\caption{The schematic illustration of how a single agent interacts with the environment. For each agent $i$ at time step $\tau$, it takes the previous belief $b^i_{\tau-1}$, previous action $a^i_{\tau-1}$ and observation $o^i_{\tau}$  into the belief network. Through differentiable transition function, likelihood function and resample function, particles and their weights are updated.  An estimated network $G$ is to estimate the approximated current state of agent $i$ (Section~\ref{Infer State}). We leverage two networks $V$ and $q$ to get the accumulated individual regret (Section~\ref{Accumulated Value Function Approximation}).  Finally, the actions are chosen from the action maximizing the accumulated regret. 
} \label{fig2}
\end{figure}

Our framework (shown in Fig.~\ref{fig1} and Fig.~\ref{fig2}) consists of four-folds: 
(1) group the agents who cooperate with each other as a \emph{team} and build a centralized training RM method for that \emph{team};
(2) learn the relationship between team regret and each individual regret by designing a shaping-form decomposition method for dividing the team regret to each agent's individual regret; 
(3) use current belief and observation to infer the current true state of each agent;
(4) design a novel training scheme to train the counter-factual value function. 


\subsection{Team Regret Minimization}\label{Team Regret Minimization}

As we mentioned above, for a cooperative setting, agents have the same goal can be treated as a team and we can leverage RM to the team. 
we named this method as team regret minimization.
And we can define the accumulated \emph{team}  regret as:
\begin{small}
\begin{equation}
\label{3}
\widehat{\operatorname{REG}}_{1:T}(\mathbf{I},\mathbf{a})=\sum\nolimits_{t=1}^{T}\left(\widehat{Q}_{\boldsymbol{\pi}_t}(\mathbf{I},\mathbf{a})-\widehat{V}_{\boldsymbol{\pi}_t}(\mathbf{I})\right),
\end{equation}
\end{small}\normalsize
where  
$\mathbf{I} = \langle I^{i}\rangle_{i=1}^{N}$ is the information state for the team at episode $T$ and $\mathbf{a} = \langle a^{i}\rangle_{i=1}^{N}$ is the action for the team. 
$\widehat{V}(\cdot)$ is the counter-factual state value function of the team while $\widehat{Q}(\cdot,\cdot)$ is the counter-factual state-action value of the team. 
For simple representation, $\widehat{Q}_{\boldsymbol{\pi}_t}$, $\widehat{V}_{\boldsymbol{\pi}_t}$, and $\widehat{\operatorname{REG}}_{1:T}(\mathbf{I},\mathbf{a})$ hereinafter referred to as $\widehat{Q}$, $\widehat{V}$ and $\widehat{\operatorname{REG}}_{1:T}$ unless stated otherwise.    

\subsection{Decomposition Methods}
This section we discuss the details of our decomposition methods, involving the relationship between team and individual regrets, the additive-form decomposition, and the more general shaping-form decomposition method.
\subsubsection{Decomposition Method for Assigning Individual Regret}
An essential issue is how to design a decomposition mechanism to obtain the individual counter-factual values and the regret based policies for each agent. 
Inspired by~\citep{nguyen2018credit,sunehag2017value,rashid2018qmix}, we propose two types of decomposition methods: an additive-form decomposition as well as a shaping-form decomposition. 
Before we discuss more details about these methods, we should first construct the \emph{consistency} between joint policy (team policy) and individual policies at $t_1$-th episode. The consistency between team policy and individual policy can be formed as: $
     \boldsymbol{\pi}(\widehat{\operatorname{REG}}_{1:t_1})
     =\left[\pi(\operatorname{REG}^{1}_{1:t_1}), {\dots}, \pi(\operatorname{REG}^{N}_{1:t_1})\right]
$, where $\pi(\cdot)$ and $\boldsymbol{\pi}(\cdot)$ are the individual and team policies to select action according to the regret (e.g., regret matching policy or greedy policy). $\operatorname{REG}^{i}_{1:t_1}=\sum^t_{t=1}(Q^i(I^i,a^i)-V^i(I^i))$ is the individual regret for agent $i$, and $Q^{i}(I^i,a^i)$ as well as $V^{i}(I^i,a^i)$ which satisfy the relationship: $V^{i}(I^i) = \sum_{a^i\in A}\pi^i(a^i|I^i)Q^{i}(I^i,a^i)$, represent counter-factual state-action value, and counter-factual state value respectively (to evaluate the performance of agent $i$ at information state $I^i$ while taking action $a^i$). 
The key idea is that the under this consistency, \emph{the joint actions selected by a team is equal to the policies individually selected by each agent in that team}. In fact, if both team and individual agents always choose action strategies that maximize the positive regrets (i.e., the regret is large than zero), we can further simplify the relationship as: 
\begin{small}
\begin{equation}
\label{consisteq}
\arg \max_{\boldsymbol{a}} (\widehat{\operatorname{REG}}_{1:t_1})_{+} = 
\left[\begin{array}{c} {\arg \max_{a^1} (\operatorname{REG}^{1}_{1:t_1}(I_\tau^1,a_\tau^1))_{+}} \\ {\vdots} \\{ \arg \max_{a^N} (\operatorname{REG}^{N}_{1:t_1}(I_\tau^N,a_\tau^N))_{+}} \end{array}\right],
\end{equation}
\end{small}\normalsize
where $(\cdot)_{+} = \max\{0,\cdot\}$ is the positive clip function. This is the $\max$ form of the \emph{consistency} between joint policy and individual policies and the following are the two practical decomposition methods satisfying this consistency. 

\textbf{Additive-form Decomposition Method}. For additive-form decomposition method, we construct an equation to support the decomposition method, which implies an idea to decompose the team value function averagely to each agent: $\widehat{Q}(\mathbf{I_\tau},\mathbf{a_\tau})= \sum\nolimits_{i=1}^N Q^i(I_\tau^i,a_\tau^i),   \widehat{V}(\mathbf{I_\tau}) = \sum\nolimits_{i=1}^N V^i(I_\tau^i)\nonumber$. This is the \emph{Additive-form Decomposition}. Based on this, we can guarantee the consistency under the additive-form team regret decomposition method:

\begin{proposition}
\label{add_ass}
If the Additive-form Decomposition holds, the consistency between joint policy and individual policies (i.e, Eq.~(\ref{consisteq})) is established. 
\end{proposition}
All the proofs can be found in Supplementary Material (\textbf{SM}).
\par Proposition \ref{add_ass} reveals that if the additive-form decomposition is guaranteed, then the joint policy is the same as the collection of individual policies of each agent. Thus, our target is to design a method to train this team regret under team reward (Details can be found in Section \ref{sec:training}). 
Now, according to Eq. (\ref{consisteq}), both the team policy (joint policy) and the individual policies can be got by:
\begin{small}
\begin{equation}
\label{22}
   \boldsymbol{\pi}(\mathbf{a}_{\tau}|\mathbf{I}_{\tau}) =                 
  \begin{bmatrix}
  \pi^1(a^1_{\tau}|I^1_{\tau}),\pi^2(a^2_{\tau}|I^2_{\tau}),\dots,\pi^N(a^N_{\tau}|I^N_{\tau})
   \end{bmatrix}, 
\end{equation}
\end{small}\normalsize
where 
$\pi^i( a_\tau^i|I_\tau^i) = \arg\max_{ a_\tau^i} \left(\operatorname{REG}^{i}_{1:t_1}(I_\tau^i,a_\tau^i)\right)_{+}$ is agent $i$'s policy. Notice that if none of the regret is larger than $0$, the action(s) will be randomly chosen. 

\textbf{Shaping-Form Decomposition Method}. The additive-form decomposition method might be a little strong. 
Therefore, to solve this problem, we relax this assumption by proposing a more general-form decomposition method. The general-form decomposition function $f$ should both satisfy the consistency relationship between joint policy and individual policies as well as provide a more flexible structure to represent the relationship between team regret and individual regrets.

In practice, to accelerate training and inference, we make a \emph{shaping-form decomposition}: The function $f$ is set as: $f(\operatorname{REG}^1_{1:t},\dots,\operatorname{REG}^N_{1:t},s)=\sum_{i=1}^N \operatorname{REG}^i_{1:t}+c(s),$ where $c(\cdot)$: $\mathcal{S}\rightarrow \mathbb{R}$ is a mapping from states to a real number value. Introducing $c(s)$ 
provides a more complicated structure to represent the relationship between team and individual agents as well as supplies extra state information to aid training. In fact, it is possible to design some complicated decomposition methods but we do not attempt them because this slows down the training scheme, especially in large-scale environment. Since $f$ includes a shaping part to adjust the decomposition function, we name it as team Value Regret Minimization with global-state function shaping (VRM-shaping). Similarly, the individual regret-based policy can be chosen according to Eq.~(\ref{22}) and under the \emph{shaping-form decomposition}, the \emph{consistency} is also established:

\begin{proposition}
If the Shaping-form Decomposition holds,  Eq.~(\ref{consisteq}) is also established. 
\end{proposition}

\subsection{Infer State by Particle Filter Network} \label{Infer State}
For a well-training agent, the more the agent knows about the current state, the better action it will take. However, due to the partially observable constraint, it is impossible for that agent to acquire the current (global) state $s$ perfectly. But this problem can be mitigated when the agent remembers all the history information it has observed or just maintains some belief states as a compression of the history information to speculate its true state~\citep{dibangoye2016optimally}. Following the same logic, we set the belief state of $i$-th agent at time step $\tau$ as $b^i_{\tau}$ and leverage particle filter network as the estimator for the distribution of current belief state to infer the true state. 

Our particle filter network is one of the Sequential Monte Carlo (SMC) methods. SMC is a Bayesian inference method widely used for establishing a relatively correct inference about the state of a probabilistic system in an uncertain environment~\citep{doucet2001introduction}. 
Specifically, in our environment, the belief states of agent $i$ can be approximated by the stack of $X$ particles with hidden state $h^x_{\tau} \in \mathcal{H}$ and its weight $w^x_{\tau}\in \mathcal{W}$: $b^i_{\tau} = \langle h^x_{\tau},w^x_{\tau}\rangle^{i}_{x=1:X}$, when $\tau>1$. And $b^i_1 = b_1$ (all the agents are sharing the same initial belief state $b_1$). The information state $I^i_\tau$ for agent $i$ can be approximated by the stack of current belief states and observation: $\langle b^i_{\tau-1}, a^i_{\tau-1}, o^i_\tau\rangle$. New particles are sampled by transition function $T^i(\cdot)$ as: $h_{\tau}^x = T^i(b^i_{{\tau}-1},a^i_{{\tau}-1},o^i_{\tau}),$ meaning that the new hidden state are drawn from the previous belief state, previous step action and current observation together. 
The updating rule for belief weight can be formed as: 
$w^x_\tau = \frac{L^x_{\tau}w^x_{{\tau}-1} }{\sum\nolimits_j L^j_{\tau}w^j_{{\tau}-1}},$ where the likelihood function $L_{\tau}^x $ is defined as: $L_\tau^x = Z^i(o^i_{\tau},h^x_{{\tau}-1})$. $Z^i(\cdot)$ is likelihood probability function guiding the variation of the particle weights update. Finally, the compressed information state at step $\tau$ is mapped from the observation $o_{\tau}^i$ and belief $b_{\tau}^i$ by state generative function as: $\kappa_{\tau}^i = G^i(o_{\tau}^i,b_{\tau}^i)$. 
To train as an end to end framework, $T^i(\cdot)$, $Z^i(\cdot)$ and $G^i(\cdot)$ are all represented as the neural networks.

\par However, as most of the particles' weights will drop to zero, there will be a particle degeneracy during the state inference~\citep{doucet2001introduction}.  
Those methods depending on the resample trick are not suitable for our end to end training. 
Following~\citep{karkus2018particle}, a soft-resample method is used to make the network differentiable: $w^x_\tau = \frac{\beta\times L^x_{\tau}w^x_{\tau-1} +(1-\beta)\times1/X }{\sum\nolimits_{j=1}^X (\beta\times L^j_{\tau}w^j_{\tau-1} ) +(1-\beta)}\nonumber$, where $\beta$ is the hyper-parameter to balance the resample ratio. The inference process of agent $i$'s particle filter at $\tau$-th step can be expressed by the function $\kappa^i_\tau = \operatorname{B}(b^i_{{\tau}-1},o_{\tau}^i)$. Details can be found in Algorithm \ref{alg3}. We name our VRM-shaping method with Bayesian inference (the particle filter) as: \emph{BVRM-shaping}.

\begin{algorithm}
  \caption{Belief Update Function}
  \label{alg3}
  \begin{algorithmic}
\STATE\textbf{Input}: $\mathbf{b}_{\tau-1}$$,\mathbf{w}_{\tau-1},\mathbf{a}_{\tau-1}$$,\mathbf{o}_\tau$
\FOR{each agent $i$}
 \STATE calculate the hidden state $h^x_\tau$ and the weights $w^x_\tau$ for beliefs through functions $T^i$, $Z^i$ and $w^x_\tau$;
  \STATE calculate each agent belief $b^i_\tau = \langle h^x_\tau,w^x_\tau\rangle_{x=1:X} $;
  \STATE calculate the compressed information state $\kappa_{\tau}^i$ through functions $G^i$;
\ENDFOR
\STATE stack the compressed information states, belief states and weights together as $\boldsymbol{\kappa}_\tau$, $\mathbf{b}_\tau$, $\mathbf{w}_\tau$;
 \STATE \textbf{RETURN} $\boldsymbol{\kappa}_\tau$, $\mathbf{b}_\tau$, $\mathbf{w}_\tau$;
  \end{algorithmic}
\end{algorithm}


\begin{figure*}[tb!]
\centering
\subfigure[The two-step matrix game. The numbers in matrices are rewards.]{
\includegraphics[width=0.30\textwidth]{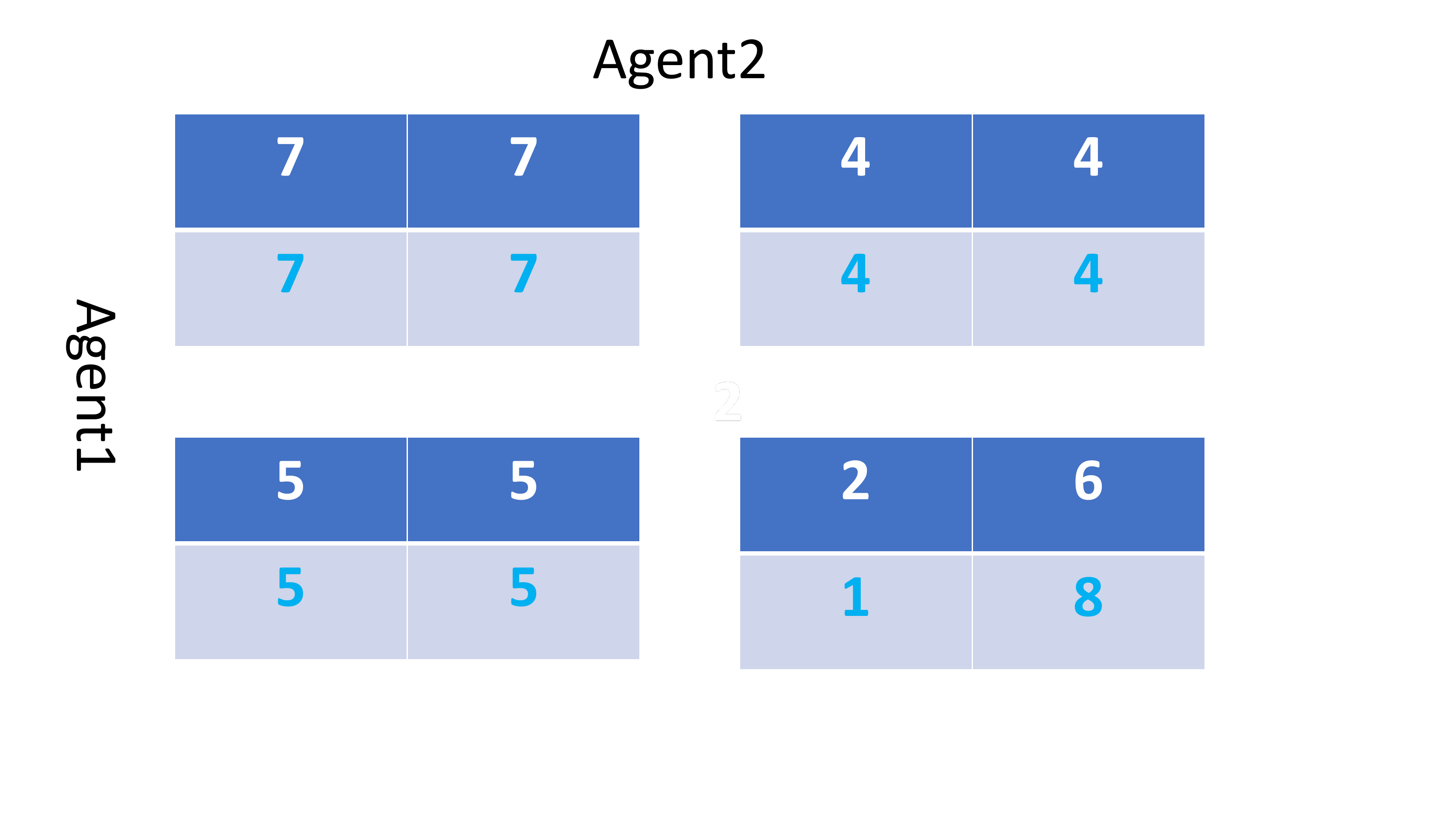}
\label{fig3a}}
\subfigure[Reward curves for methods comparisons.]{
\includegraphics[width=0.30\textwidth]{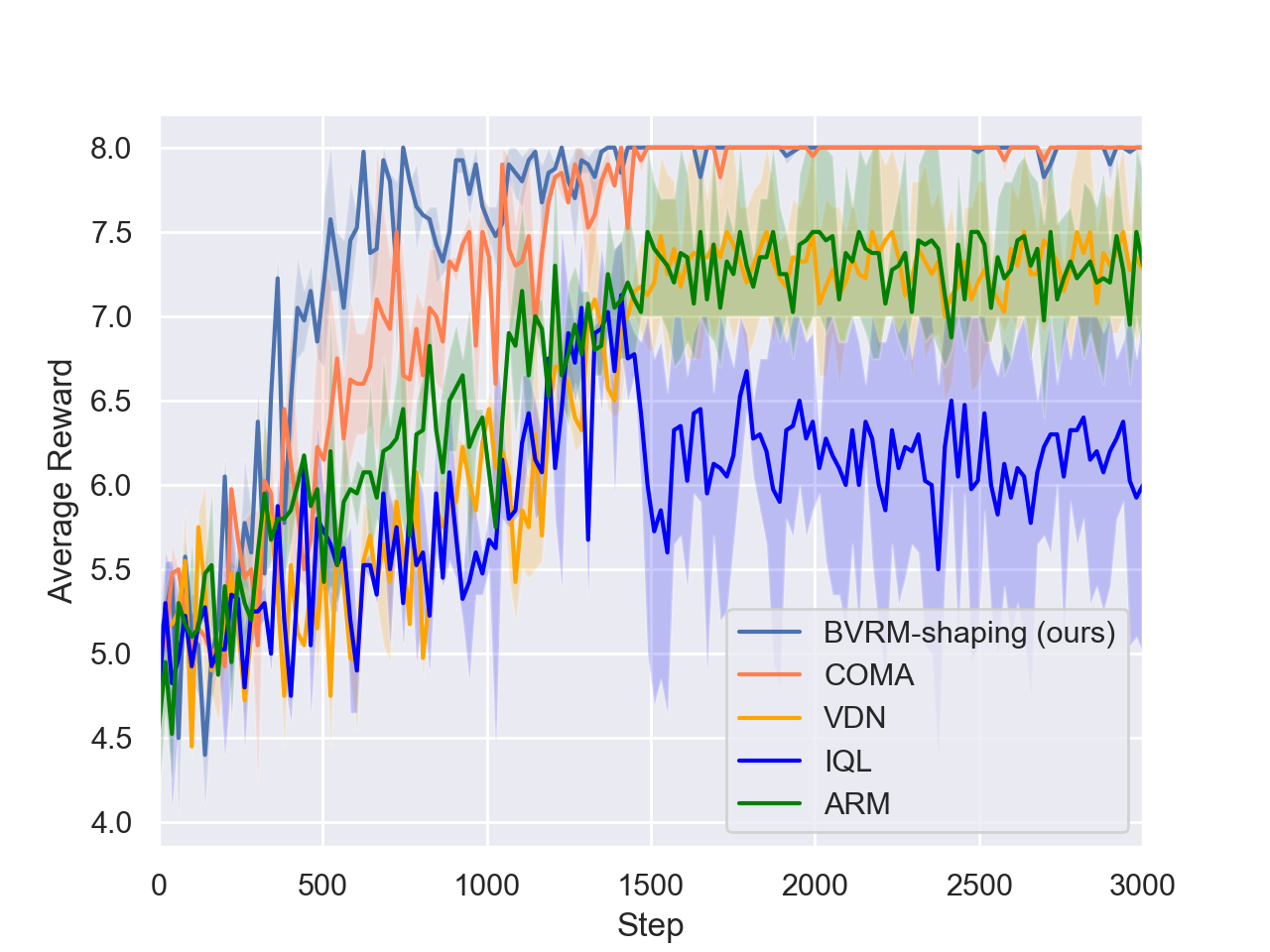}
\label{fig3b}}
\subfigure[Reward curves for ablations.]{
\includegraphics[width=0.30\textwidth]{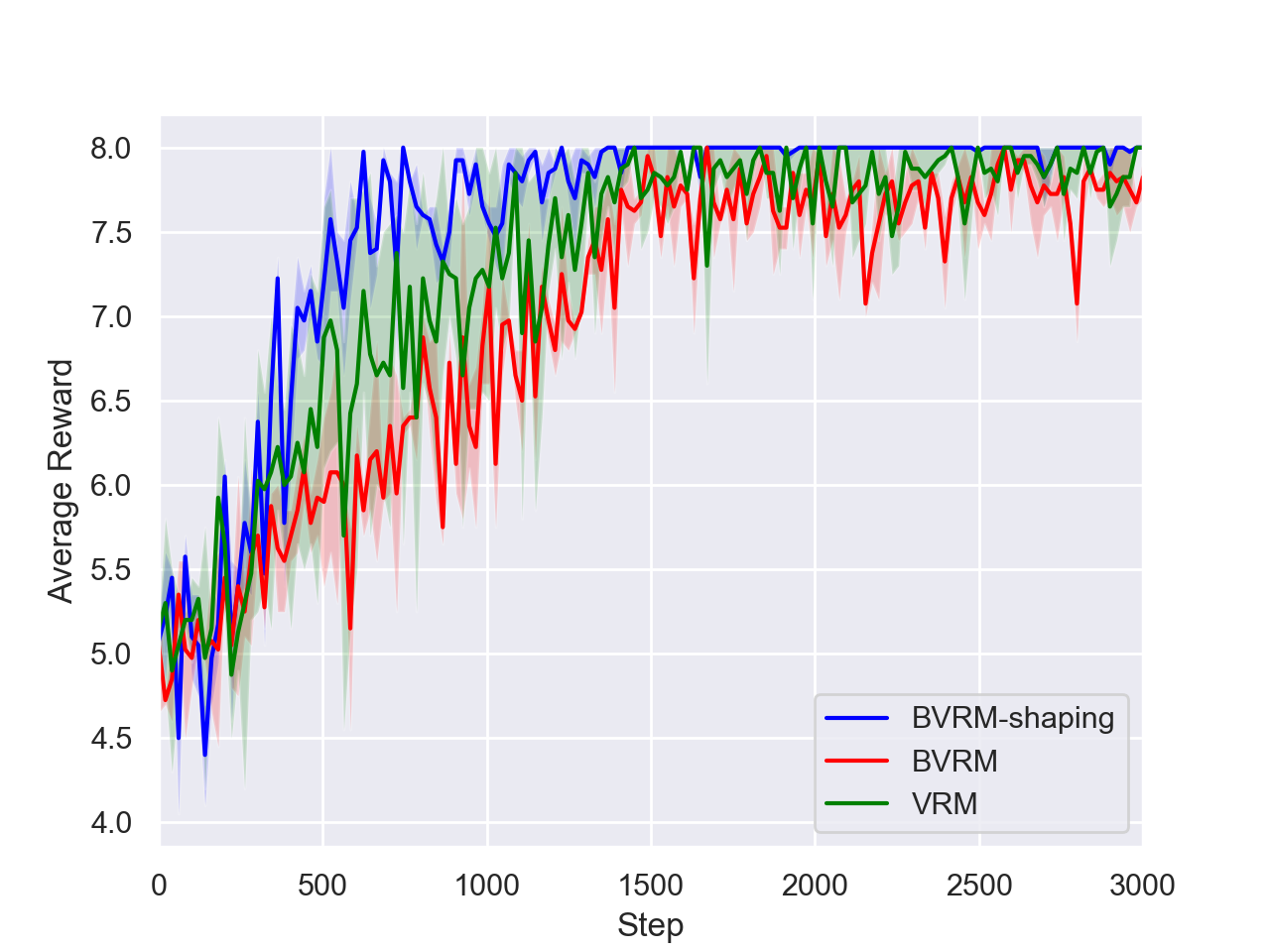}
\label{fig3c}}
\caption{The learning curve of different algorithms in the Matrix Games (recorded every 10 rounds of training with self-play).}
\label{fig:matrix_train}
\end{figure*}

\subsection{Training}
\label{sec:training}
\subsubsection{Accumulated Value Function Approximation.}\label{Accumulated Value Function Approximation}
To implement our method into practice, we are still facing one crucial problem: It is time-consuming and memory-wasting to implement Eq. (\ref{3}) because this method is required to memorize all the pairs of information states and corresponding counter-factual values, which is intractable in large-scale environment. Therefore, we need to find an alternative approach to calculate the team regret. Here, notice that:
\begin{small}
\begin{align*}
\widehat{\operatorname{REG}}_{1:t+1}(\mathbf{I},\mathbf{a})= \widehat{\operatorname{REG}}_{1:t}(\mathbf{I},\mathbf{a})
&+\widehat{Q}_{\boldsymbol{\pi}_{t+1}}(\mathbf{I},\mathbf{a})-\widehat{V}_{\boldsymbol{\pi}_{t+1}}(\mathbf{I}),
\end{align*}
\end{small}\normalsize
where $\widehat{\operatorname{REG}}_{1:t}(\mathbf{I},\mathbf{a})$ 
is the accumulated team regret at $t$ episode. $\boldsymbol{\pi}_{t+1}$ is to emphasize the joint policy at $t+1$ episode. The equation above indicates that the regret function can be updated recursively. Based on it, for shaping-form decomposition, we define the accumulated $t$ episode single agent state-value and value function as: $Q^i_{1:t}(I^i,a^i)=\sum_{t^{\prime} =1}^{t}Q^i_{t^{\prime}}(I^i,a^i)$ and $V^i_{1:t}(I^i)=\sum_{t^{\prime} =1}^{t}V^i_{t^{\prime}}(I^i,a^i)$. Based on the shaping form decomposition relationship, we have: 
\begin{small}
\begin{align*}
\widehat{\operatorname{REG}}_{1:t}(\mathbf{I},\mathbf{a})&=\sum\nolimits_i Q^i_{1:t}(I^i,a^i) - \sum\nolimits_i V^i_{1:t-1}(I^i)-\\&\sum\nolimits_i V^i_t(I^i)+  \sum\nolimits^t_{t^\prime=1} c_{t^\prime}(s).
\end{align*}
\end{small}\normalsize
Here, we  use neural networks to approximate $ Q^i_{1:t}(I^i,a^i) -  V^i_{1:t-1}(I^i)$ by $q^i_{1:t}(I^i,a^i;\omega^i)$, $V^i_{t}(I^i,a^i)$ by $V^i_{t}(I^i,a^i;\theta^i)$, and $ \sum\nolimits^t_{t^\prime=1} c_{t^\prime}(s)$ by $f(s;\xi)$, where $\theta^i$, $\omega^i$ are the parameters of neural networks for agent $i$ and $\xi$ is the parameter for $f$, the state shaping function. 
Moreover, if we neglect $f(s;\xi)$, that is exact the additive-form decomposition method. Now, through this approximation, we can leverage the neural networks to infer the counter-factual values rather than memorizing all the information states and corresponding team counter-factual values pair.
    
    
\subsubsection{Loss Functions.} Here, we describe how to train our \emph{BVRM-shaping} method under the accumulated value function approximation. In order to accelerate the training process, all the agents share the same parameters $\omega$ and $\theta$. Target network $V^i_{target}(\cdot;\theta^{'})$ is introduced to stabilize the training process~\citep{mnih2015human}, where $\theta^{'}$ is the trainable parameters. Following the same logic of
~\citep{jin2017regret}, the parameters for $q$ network $\omega$,$V$ network $\theta$, state shaping network $\xi$, and particle filter network $\lambda$, can be optimized jointly. The objective function of $q$ network in $\tau_0$-th step $t$-th episode is:
\begin{small}
\begin{align}
    \label{40}
    J_q^{\omega} &= \mathop{\mathbb{E}}
    \bigg[\frac{1}{2}(\sum\nolimits_{i=1}^Nq_{1:t}^i(I_{\tau_0}^i,a_{\tau_0}^i;\omega) \nonumber
    -\sum\nolimits_{i=1}^N q_{1:t-1}^i(I_{\tau_0}^i,a_{\tau_0}^i;\omega) -  \\& \sum\nolimits_{i=1}^N \gamma V^i_{target}(I_{\tau_0+1}^i;\theta^{'})
    - r_{\tau_0})^2 \bigg].
\end{align}
\end{small}\normalsize
For the accumulated value functions, we have:
\begin{small}
\begin{align}
\label{50}
J_V^{\theta} &= \mathop{\mathbb{E}}
\bigg[\frac{1}{2}(\sum\nolimits_{i=1}^N \mathbb{A}^{i}(k,\tau_0) -\sum\nolimits_{i=1}^N V_{t}^i(I_{\tau_0}^i;\theta)+ f(s_{\tau_0};\xi))^2\bigg],
\end{align}
\end{small}\normalsize
where $\mathbb{A}^{i}(k,\tau_0)$ is the $k$-step advantage function at time step $\tau$ for agent $i$:
\begin{small}
\begin{equation*}
    \mathbb{A}^{i}(k,\tau_0) = \bigg(\sum^{k+\tau_0}_{\tau=\tau_0} \gamma^{\tau-{\tau_0}} r^i_\tau\bigg) + \gamma^{k+\tau_0+1} V^i_{target}(I^i_{k+\tau_0+1};\theta^{'}).
\end{equation*}
\end{small}\normalsize
According to the derivation scheme, we have the gradient of the parameters $\xi$ and $\lambda$:
\begin{small}
\begin{align*}
\nabla_{\xi}J_C^{\xi} &=  \mathop{\mathbb{E}}
[\sum_{i=1}^N\mathbb{A}^{i}(k,\tau_0)-\sum_{i=1}^N V_{t}^i(I_{\tau_0}^i;\theta)+ f(s_{\tau_0};\xi))\nabla_{\xi}f(s_{\tau_0};\xi)], \\
\nabla_{\lambda}J_b^\lambda &=  \mathop{\mathbb{E}}
[\frac{\partial J_V^{\theta}}{\partial \kappa_{\tau_0}^i}\nabla_{\lambda}B(b_{{\tau_0}-1}^i,o_{\tau_0}^i;\lambda)].
\end{align*}
\end{small}\normalsize
The differentiation of $J_C^{\xi}$ and $J_b^\lambda$ are both derived from Eq. (\ref{50}). 

All the parameters are updated by stochastic gradient descent. We want to stress that the updating methods of $q$ and $V$ are similar to temporal-difference methods~\citep{sutton1998introduction} but differently, our $q-V$ represents the \emph{accumulated counter-factual value} rather than one episode advantage value.

\emph{Due to the space limitation, details about the losses and the pseudocode can be found in \textbf{SM} B and C.} 

\section{Experimental Evaluation}\label{exp}
We evaluate and analyze our methods in two widely used scenarios: the cooperative two-step matrix game and the mixed cooperative-competitive battle game. The cooperative two-step matrix game is a toy environment while the mixed cooperative-competitive battle game is \emph{TONE}, especially when the number of agents is large. We conduct many comparisons and ablations in our experiments. For comparisons, we compare Independent Q-Learning: \emph{IQL}~\citep{tan1993multi}; individual regret minimization methods: \emph{ARM}\footnote{\emph{ARM} is the individual regret minimization methods, i.e., each agent uses the RM independently using the team reward.} ~\citep{jin2017regret}; the Counter-factual Multi-Agent Policy Gradient:  \emph{COMA}~\citep{foerster2018counterfactual}; and Value Decomposition Networks: \emph{VDN}~\citep{sunehag2017value}. 

For ablations, 
firstly, we focus on the centralized team value regret minimization with additive-form decomposition, named \emph{VRM}. Secondly, we consider the combinations of particle filter network (Bayesian inference) and VRM as \emph{BVRM}. We also add global-state shaping into BVRM as \emph{BVRM-shaping} (the shaping-form decomposition). 
\subsection{The Matrix Game}

\textbf{Environment:} To study the effectiveness of our algorithms, we leverage a two-agent, two-step cooperative matrix game: in the first step, each agent chooses a matrix (out of four) independently.
The payoff of each agent is initialized as zero. Afterwards, each agent chooses the location of the step based on the matrix chosen in step one as shown in Fig.~\ref{fig3a}. Notice that in this step, the agents do not have the knowledge of the others' choices and finally all the agents will receive a global payoff (reward) $r = [column][row][subcolumn][subrow]$. 

\par
\textbf{Results:} \emph{For comparison results}, Fig.~\ref{fig3b} implies that \emph{BVRM-shaping} outperforms other methods dramatically both in final reward value and convergence speed, implying that our method performs well in the two-step matrix game. 
Moreover, \emph{COMA} is better than \emph{VDN}, indicating that the counter-factual method might be advantageous over the traditional Q-learning but still worse than our \emph{BVRM-shaping} method, which testifies the advantage of team RM. \emph{For ablations results}, all our methods have the ability to find the optimal solution. It indicates that our methods can perform well in the two-step matrix game. 
There is a very interesting phenomenon in Fig.~\ref{fig3c} is that the convergence speed of \emph{VRM} is faster than that of \emph{BVRM}. This may be caused by the small number of the states where the particle filter network is unnecessarily needed in that environment (only two steps). \emph{The details of the environment, the tableau results, the hyper-parameters and the structure of networks can be found in \textbf{SM} D}.

\begin{figure*}[tb!]
	\centering
	\subfigure[the win-rate for method comparisons in the 64-agent battle game.]{
		\includegraphics[width=0.31\textwidth]{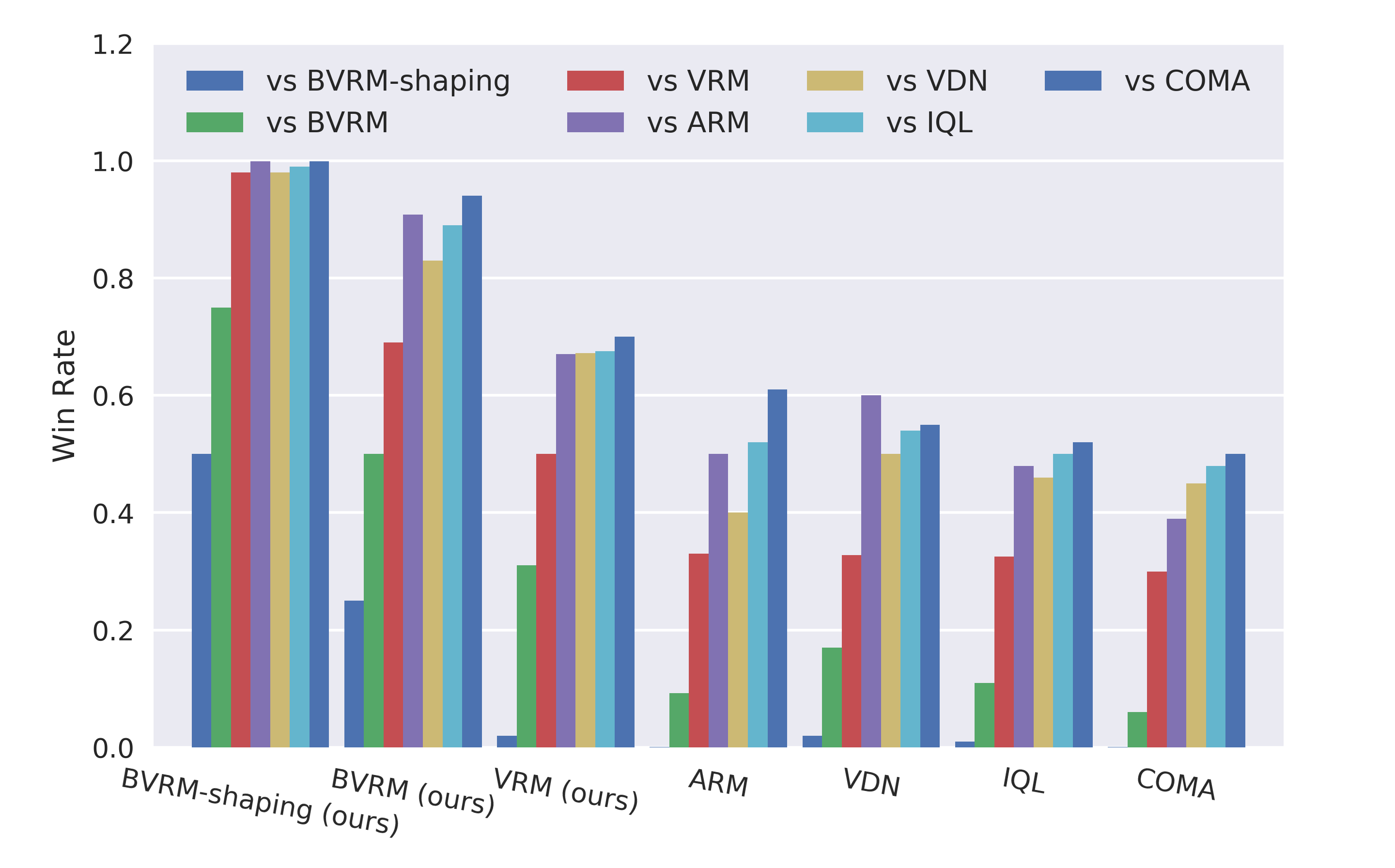}
		\label{fig4a}}
	\subfigure[the win-rate for method comparisons in the 256-agent battle game.]{
		\includegraphics[width=0.31\textwidth]{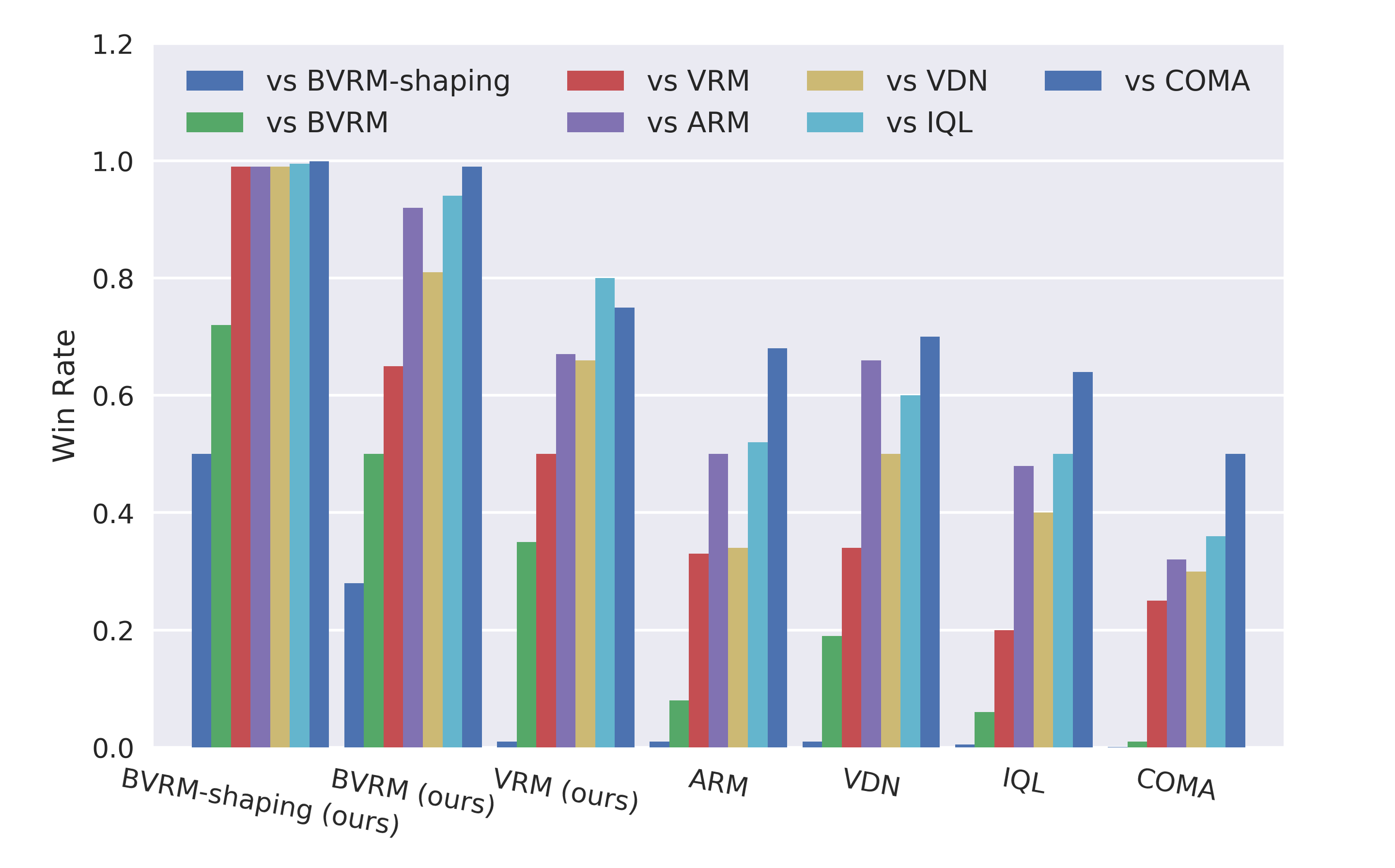}
		\label{fig4b}}
	\subfigure[The win-rates change in environment from 64 agents to 256 agents.]{
		\includegraphics[width=0.31\textwidth]{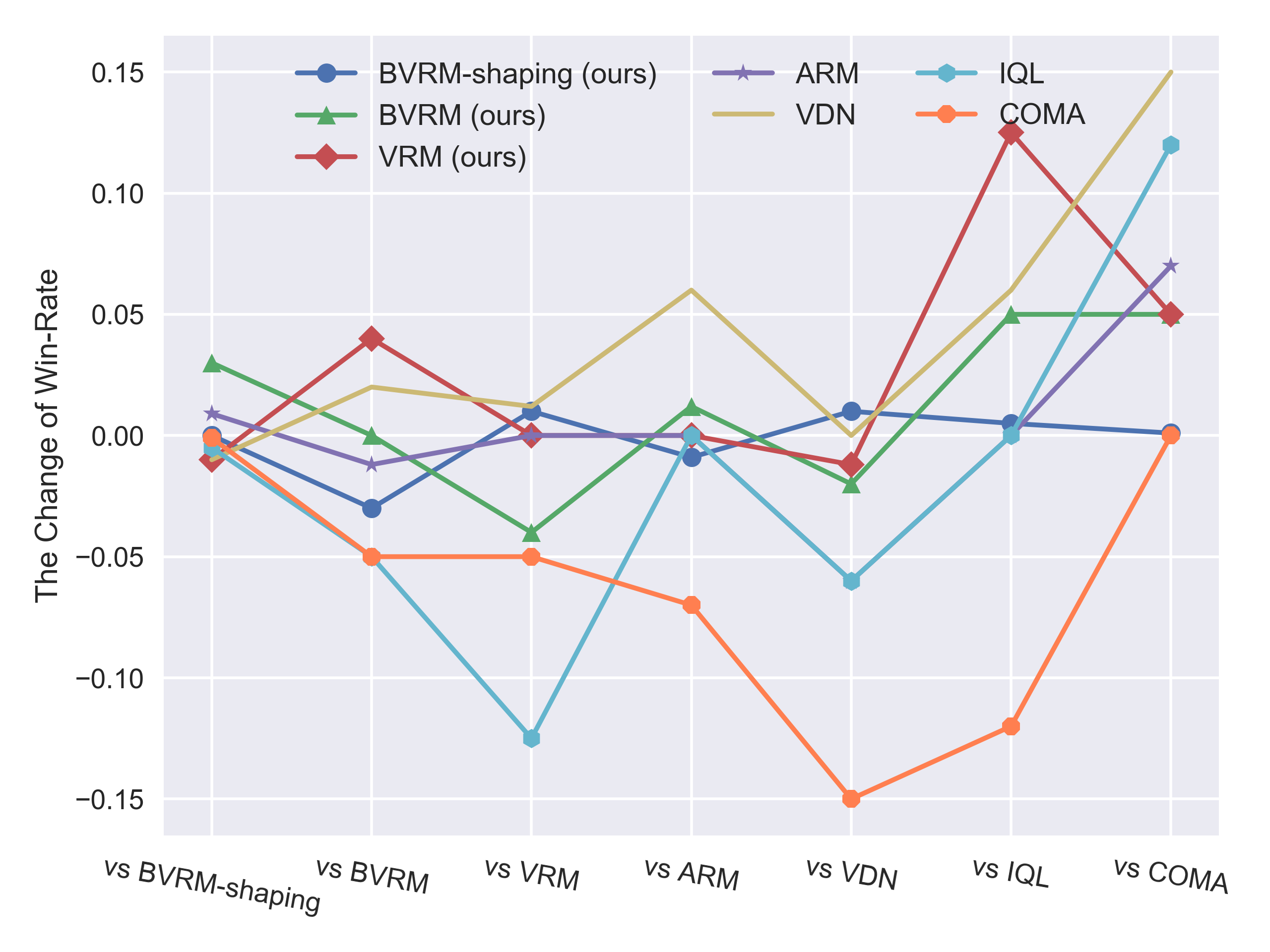}
		\label{fig4c}}
	\caption{The learning curve of different algorithms in the Battle Game. Figures (a) and (b) are the win-rate for method comparisons in the 64 and 256 agent battle game respectively. Figure (c) implies the win-rates change in environment from 64 agents to 256 agents and the points below $y$-axis mean the corresponding methods perform worse in 256 agent environment than in 64 agent environment, and vice versa. All the results shown on the figures (a), (b), (c) are the average of 50 rounds battle game for each experiment.}
	\label{fig:battle_train}
\end{figure*}

\subsection{The Battle Game}
\par
\textbf{Environment:}
The second study is undertaken on a large-scale multi-agent system called Magent~\citep{zheng2018magent}, a mixed cooperative-competitive game that simulates the battle between two armies. In this system, each agent aims to cooperate with their teammates to destroy their opponents. Agent can take actions to either move to or attack nearby grids. The default settings are applied in our experimental study. \emph{In order to evaluate the scalability of our approaches, two battle scenarios are introduced here: the first one consists of 64 agents while the other one consists of 256 agents.} The rewards are the sum of the action utilities of each agent\footnote{The original reward is the individual reward. To make it suit our team reward assumption, we modify the environment that each agent can only get access to the sum of each individual reward.}. The utilities of actions are defined as: moving one step, killing an enemy, attacking an enemy, attacking a blank grid, and being attacked or killed are -0.005, 5, 0.2, -0.1 and -0.1.  
\begin{figure*}
  \centering
   \centerline{		\includegraphics[width=0.95\linewidth]{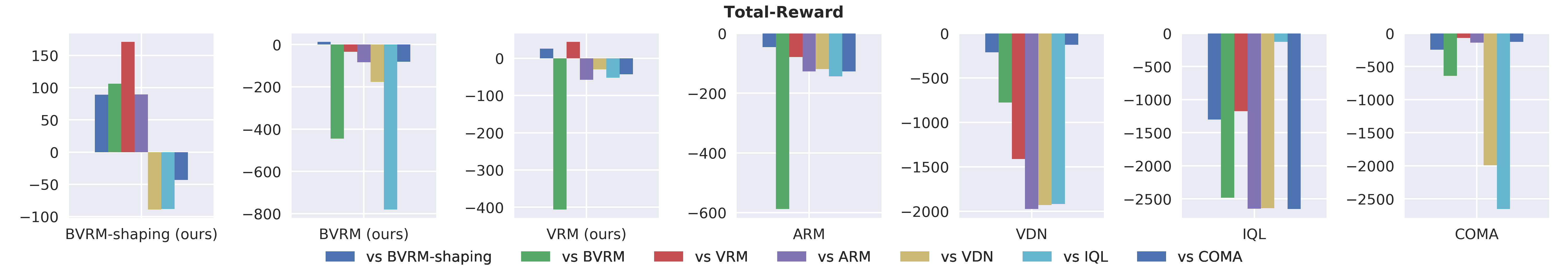}}
\caption{The average total reward for method comparisons in the 64 agent battle game. The results shown on the figure are the average of 50 rounds battle game for each comparative experiment.} \label{fig5a}
\end{figure*}

\textbf{Results:}
From Fig.~\ref{fig:battle_train} we conclude that our methods outperform other state-of-the-arts in all scenarios. This means that our approaches are both well-performing and scalable in complicated (TONE) environments. In fact, \emph{ARM} performs poorly than all of our methods, testifying the assumption that the individual regret minimization method is hard to  induce cooperation. 
\emph{For comparison,} \emph{ARM} is an independent training method but it reaches a better score than \emph{IQL}. This result presents a view that the independent RM training scheme might be better than the independent Q-learning in mixed cooperative and competitive environment. 
Comparing with \emph{COMA}, which is hard to train when the number of agents increases, our approaches are consistent with high win-rate in all the scenarios as shown in Fig.~\ref{fig4c}, supporting that our method can be adopted to various environments. Moreover, as in Fig.~\ref{fig5a} 
, the reward for \emph{BVRM-shaping} is higher than other state-of-the-art methods, representing that our method can learn a better strategy than other methods. 
\emph{For ablations,} from Fig.~\ref{fig:battle_train} and Fig.~\ref{fig5a}, 
 we summarize that the rank of the performance is \emph{BVRM-shaping} $>$ \emph{BVRM} $>$ \emph{VRM}. Specifically, comparing with the method with state-shaping, (\emph{BVRM-shaping}) and without (\emph{BVRM}), we find that the method with global-state shaping could achieve a significant improvement. This reveals the significance of finding the correct decomposition methods for the value function. There is also an interesting phenomenon in Fig.~\ref{fig5a} where comparing with \emph{BVRM-shaping} and \emph{BVRM}, \emph{BVRM-shaping} has a high positive average total reward while \emph{BVRM} has a low negative average total reward. This occurs due to the learning strategies they have learned: the  \emph{BVRM-shaping} learns an offensive strategy while \emph{BVRM} learns a relatively defensive strategy. More details (including the average total rewards for $256$ agents) about the environment and the hyper-parameters are in \textbf{SM} E.


\section{Conclusion}

This paper proposes a team RM based Bayesian MARL framework to solve the team partially observable or non-stationary environment. 
Specifically, our method introduces a novel team regret training scheme to obtain a well-performing team (meta agent) policy as well as a novel decomposition method to decompose the team regret to generate the policy for each agent in decentralized execution. To further improve the performance of our method, we design a novel end to end particle network to provide a better inference of current state. The experiments in the two-step matrix game (cooperative game) and large-scale battle games (mixed cooperative-competitive game) reveal that our methods can significantly outperform other baselines in various environments. 

In the future, we will introduce how to model the opponents' behaviours into our frameworks. We are also willing to test our methods on solving real-world practical problems. \emph{Analysis, experiment details, code, and video can be found in supplementary materials.}


\bibliographystyle{aaai}
\bibliography{aaai2020}

\end{document}